%% file: main.tex
\documentclass[10pt,twocolumn,letterpaper]{article}

\usepackage[pagenumbers]{wacv} 

\input{preamble}

\definecolor{wacvblue}{rgb}{0.21,0.49,0.74}
\usepackage[pagebackref,breaklinks,colorlinks,allcolors=wacvblue]{hyperref}

\def\wacvPaperID{657}
\def\confName{WACV}
\def\confYear{2026}

\title{STRinGS: Selective Text Refinement in Gaussian Splatting
\vspace{-3mm}}

\author{
Abhinav Raundhal\textsuperscript{*} \quad
Gaurav Behera\textsuperscript{*} \\
P. J. Narayanan \quad
Ravi Kiran Sarvadevabhatla \quad
Makarand Tapaswi \\ 
{\normalsize CVIT, IIIT Hyderabad, India} \\
{\texttt{\normalsize \href{https://strings-official.github.io/}{STRinGS-official.github.io}}}
}

\begin{document}

\twocolumn[{
\maketitle
\input{figures/teaser}

}]

{
  \renewcommand{\thefootnote}{\fnsymbol{footnote}}
  \footnotetext[1]{Equal contribution}
}

\input{sec/0_abstract}
\input{sec/1_intro}

\input{sec/2_related_work}
\input{sec/3_dataset}

\input{sec/4_methodology}

\input{sec/5_exp_and_results}

\input{sec/6_conclusion}
\newpage
{
\small
\bibliographystyle{ieeenat_fullname}
\bibliography{bib/longstrings, bib/main}
}

\end{document}

%% file: preamble.tex
\usepackage{booktabs, multirow, xcolor, colortbl}
\usepackage{array}
\usepackage{tabularx}
\definecolor{first}{RGB}{242, 182, 181}
\definecolor{second}{RGB}{249, 218, 184}
\definecolor{third}{RGB}{254, 250, 199}
\definecolor{gray}{RGB}{225, 225, 225}
\usepackage{array, makecell}
\usepackage[ruled, vlined]{algorithm2e}
\usepackage{pifont}
\usepackage{soul}
\usepackage[normalem]{ulem}
\usepackage{float}

\newcommand{\hlfirst}[1]{\sethlcolor{first}\hl{#1}}
\newcommand{\hlsecond}[1]{\sethlcolor{second}\hl{#1}}
\newcommand{\hlthird}[1]{\sethlcolor{third}\hl{#1}}

\newcommand{\comments}[1]{}

\newcommand{\methodfullname}{Selective Text Refinement in Gaussian Splatting}
\newcommand{\methodname}{STRinGS}
\newcommand{\dataname}{STRinGS-360}
\renewcommand{\paragraph}[1]{\vspace{1mm}\noindent\textbf{#1}}

\definecolor{confgray}{RGB}{128, 128, 128}
\newcommand{\papertable}[1]{{\tiny \color{confgray} {#1}}}

\newcommand{\mcV}{\mathcal{V}}

\newcommand{\mcM}{\mathcal{M}}
\newcommand{\mcP}{\mathcal{P}}
\newcommand{\Ptext}{\mathcal{P}_{\text{text}}}
\newcommand{\Pnontext}{\mathcal{P}_{\text{non-text}}}
\newcommand{\Gtext}{\mathcal{G}_{\text{text}}}
\newcommand{\Gnontext}{\mathcal{G}_{\text{non-text}}}
\newcommand{\Gtextrefined}{\Gtext^{\text{refined}}}
\newcommand{\bfP}{\mathbf{P}}

\newcommand{\bfg}{\mathbf{g}}
\newcommand{\Nmax}{N_{\text{max}}}

\newcommand{\mcL}{\mathcal{L}}

\renewcommand{\Return}[1]{\textbf{Return} #1}

%% file: figures/teaser.tex
\begin{center}
\vspace{-6mm}
\includegraphics[width=\linewidth]{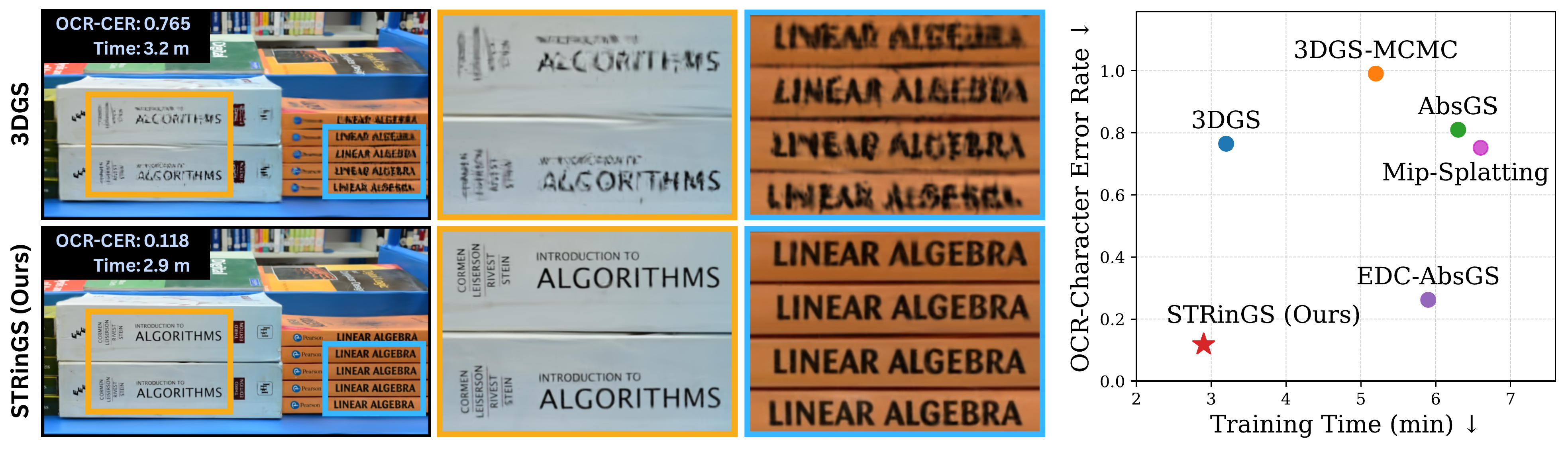}
\vspace{-6mm}
\captionof{figure}{
Qualitative and quantitative comparison of Gaussian Splatting methods on text reconstruction at 7K iterations.
\textbf{Left:}
On a novel view from the \emph{Shelf} dataset that features library books on a shelf, our approach \methodname{} (bottom) produces sharper and readable text as compared to vanilla 3DGS (top).
\textbf{Right:}
We quantify text reconstruction using Character Error Rate (CER) used in Optical Character Recognition (OCR).
The accompanying scatter plot presents readability (CER, lower is better) \vs~training time.
\methodname{} achieves the best performance both in terms of lowest error and fastest training time.
}
\vspace{2mm}
\label{fig:teaser}
\end{center}

%% file: sec/0_abstract.tex
\begin{abstract}
\label{sec:abstract}

Text as signs, labels, or instructions is a critical element of real-world scenes as they can convey important contextual information. 
3D representations such as 3D Gaussian Splatting (3DGS) struggle to preserve fine-grained text details, while achieving high visual fidelity. Small errors in textual element reconstruction can lead to significant semantic loss.
We propose \methodname{}, a text-aware, selective refinement framework to address this issue for 3DGS reconstruction. Our method treats text and non-text regions separately, refining text regions first and merging them with non-text regions later for full-scene optimization.
\methodname{} produces sharp, readable text even in challenging configurations. We introduce a text readability measure OCR Character Error Rate (CER) to evaluate the efficacy on text regions. \methodname{} results in a 63.6\% relative improvement over 3DGS at just 7K iterations. We also introduce a curated dataset \dataname{} with diverse text scenarios to evaluate text readability in 3D reconstruction. 
Our method and dataset together push the boundaries of 3D scene understanding in text-rich environments, paving the way for more robust text-aware reconstruction methods.
\end{abstract}

%% file: sec/1_intro.tex
\section{Introduction}
\label{sec:intro}

Capturing 3D scenes from multi-view images for reconstruction and novel view generation is an important problem with applications in mixed reality, robotics, entertainment, archaeology and beyond. Early methods that used explicit geometry~\cite{narayanan1998constructing} were tedious. After this, neural scene representations such as NeRF (Neural Radiance Fields) and its variants~\cite{mildenhall2021nerf,barron2023zip,muller2022instant} dominated the field.
More recently, 3D Gaussian Splatting (3DGS)~\cite{kerbl20233d} was proposed that uses a geometry-neural hybrid representation. 3DGS also achieved real-time novel-view rendering with state-of-the-art visual fidelity.

3DGS represents scenes using 3D Gaussians and progressively optimizes them, using a coarse-to-fine strategy. This strategy often struggles with high-frequency details such as in fine textured regions and text present in the scene. In particular, many real-world scenes contain text in different ways that are useful for downstream applications. For example, in autonomous navigation, text is essential for interpreting road signs and waypoint recognition, while in VR, clear text improves user experience, and in robotics, it aids object identification and manipulation. 
\cref{fig:teaser} (top) shows the low quality of text reconstructed using 3DGS.

\textit{Can 3DGS be given a pair of reading glasses to enhance visual quality and readability of text regions in the scene?}
We address this problem in this paper.
We present \methodfullname{} (\methodname{}), a novel framework for improving text readability in 3DGS reconstructions. Prior related approaches attempted to enhance high-frequency regions~\cite{ye2024absgs, deng2024efficient} or improve texture detail~\cite{rong2025gstex, xu2024texture, chao2025textured}. \methodname{} identifies text regions and selectively refines them following a two-phase strategy (\cref{sec:method}):
(i)~Phase 1 isolates text regions and selectively reconstructs them; and
(ii)~Phase 2 performs a global scene refinement that maintains background fidelity while preserving improved text quality.

Standard 3D reconstruction datasets \cite{barron2022mip, knapitsch2017tanks, hedman2018deep, jensen2014large, ling2024dl3dv} contain sparse or no text, limiting their use for evaluating our approach. We introduce \dataname{}, a curated dataset of \textit{five} text-rich 3D scenes (\cref{sec:strings_360_dataset}) to address this. Traditional image fidelity based evaluation metrics (\eg~PSNR) are also insufficient to evaluate text readability. We introduce OCR Character Error Rate (OCR-CER) as a text readability measure to compare rendered and ground-truth images using a standard Optical Character Recongizer~\cite{googlevisionapi}. \methodname{} achieves an average of 23.0\% relative improvement in OCR-CER over standard 3DGS~\cite{kerbl20233d} at 30K iterations and 63.6\% relative improvement in OCR-CER at 7K training iterations.
\cref{fig:teaser} shows the qualitative and quantitative improvement in text readability for a novel view at 7K iterations of training with \methodname{}.

The key contributions of our work are given below.
\begin{enumerate}
\item We propose \methodname{}, the first framework for explicit text refinement in 3DGS, enabling accurate and readable text in rendered novel views.
\item We introduce \dataname{}, a curated benchmark to evaluate 3D reconstruction methods on text-rich scenes
and propose OCR-CER to quantify text readability.
\item We demonstrate that \methodname{} enables superior text readability without compromising image quality compared to existing high-frequency enhancement or densification strategies. Furthermore, this is achieved in early stages of training, a critical requirement for time-constrained applications.
\end{enumerate}

%% file: sec/2_related_work.tex
\section{Related Work}
\label{sec:related_works}

Traditional 3D reconstruction uses Structure-from-Motion (SfM)~\cite{schoenberger2016sfm} and Multi-View Stereo (MVS)~\cite{schoenberger2016mvs} pipelines to recover camera poses and sparse point clouds from input images.
Neural Radiance Fields (NeRFs)~\cite{mildenhall2021nerf, gao2022nerf} from the last few years are a paradigm shift as they represent scenes as volumetric fields using MLPs, enabling photorealistic novel view synthesis at the cost of slow training.
While methods like Instant-NGP~\cite{muller2022instant} improve rendering speed, real-time rendering remains challenging.
3D Gaussian Splatting (3DGS)~\cite{kerbl20233d} addresses this by adopting anisotropic 3D Gaussians to represent 3D scenes that enable fast differentiable rasterization.
However, 3DGS struggles to preserve high-frequency details, as the coarse-to-fine optimization favors global fidelity over local structure.

\paragraph{3DGS improvements.}
Recent works extend 3DGS to improve overall scene reconstruction quality and address these limitations.
Mip-Splatting~\cite{yu2024mip} tackles aliasing and scale inconsistencies by introducing filters that make 3DGS more robust across zoom levels.
3DGS-MCMC~\cite{kheradmand20243d} introduces a sampling-based formulation to improve Gaussian initialization,
while AbsGS~\cite{ye2024absgs} addresses the over-reconstruction of fine structures by revising the gradient-based densification strategy.
Mini-Splatting~\cite{fang2024mini, fang2024mini2} proposes guided densification and simplification pipelines that maintain scene fidelity with fewer primitives.
Efficient Density Control (EDC)~\cite{deng2024efficient} is a plug-and-play module that enhances various 3DGS variants~\cite{fang2024mini, mallick2024taming, ye2024absgs} by incorporating targeted pruning and splitting operations to improve scene fidelity and efficiency.
Several other approaches densify Gaussians across the scene based on visibility, reconstruction error, or color cues to improve fidelity in detail-rich areas~\cite{chan2024point, rota2024revising, kim2024color, zhang2024pixel}.

\paragraph{Extensions to texture.}
To address the limited expressivity of standard Gaussians, \textit{texture-based extensions} have also emerged.
GSTex~\cite{rong2025gstex} and HDGS~\cite{song2024hdgs} augment 2D Gaussian splatting~\cite{huang20242d} by attaching learnable texture maps to each primitive.
Texture-GS~\cite{xu2024texture} and Textured Gaussians~\cite{chao2025textured} extend this paradigm to 3DGS, enabling better disentanglement of geometry and appearance.
Textured-GS~\cite{huang2024textured} further enhances this with spherical harmonics for spatially-varying color and opacity.
Billboard Splatting~\cite{svitov2024billboard} proposes a new representation using textured planar primitives, offering improved quality at the cost of increased training time.

\paragraph{\methodname{} focuses on text.}
While these works enhance overall visual fidelity, they do not explicitly target semantic regions such as text, which are vital for downstream applications.
In contrast, our method introduces selective refinement for text regions in 3DGS.
By decoupling the optimization of text and non-text regions, \methodname{} achieves sharper and more readable textual content with fewer training iterations and without degrading overall scene quality.

%% file: sec/3_dataset.tex
\section{\dataname{} Dataset}
\label{sec:strings_360_dataset}

\input{figures/dataset}

Existing 3D scene datasets often lack semantically meaningful text, \ie, text that provides information relevant to the scene, on foreground objects.
When present, text is typically sparse and relegated to the background, making these datasets unsuitable for evaluating methods that target text-specific refinement.
Moreover, datasets such as DL3DV-10K Benchmark~\cite{ling2024dl3dv} offer only flat or panned views rather than full 360° coverage, restricting the ability to assess text reconstruction across diverse viewpoints.

To address these limitations, we introduce \dataname{}, a curated dataset of \textit{five} indoor scenes designed to benchmark text readability in 3D Gaussian Splatting (\cref{fig:dataset}).
Each scene centers on a single or a set of object(s) containing dense, semantically meaningful text exhibiting several challenges.
A. \textit{Extinguisher} features instructional text on a curved cylindrical surface;
B. \textit{Books} contains flat, densely packed book titles with author names;
C. \textit{Chemicals} presents chemical compositions on labeled bottles in a laboratory shelf;
D. \textit{Globe} includes geographical names on a spherical surface; and
E. \textit{Shelf} shows stacks of academic books in a structured and sometimes occluded setting, with repeated titles commonly found in libraries.
These scenes span flat, cylindrical, and spherical configurations and offer a diverse and realistic benchmark for evaluating fine-grained textual fidelity in 3D reconstructions.

%% file: figures/dataset.tex
\begin{figure}[t]
\centering
\includegraphics[width=\linewidth]{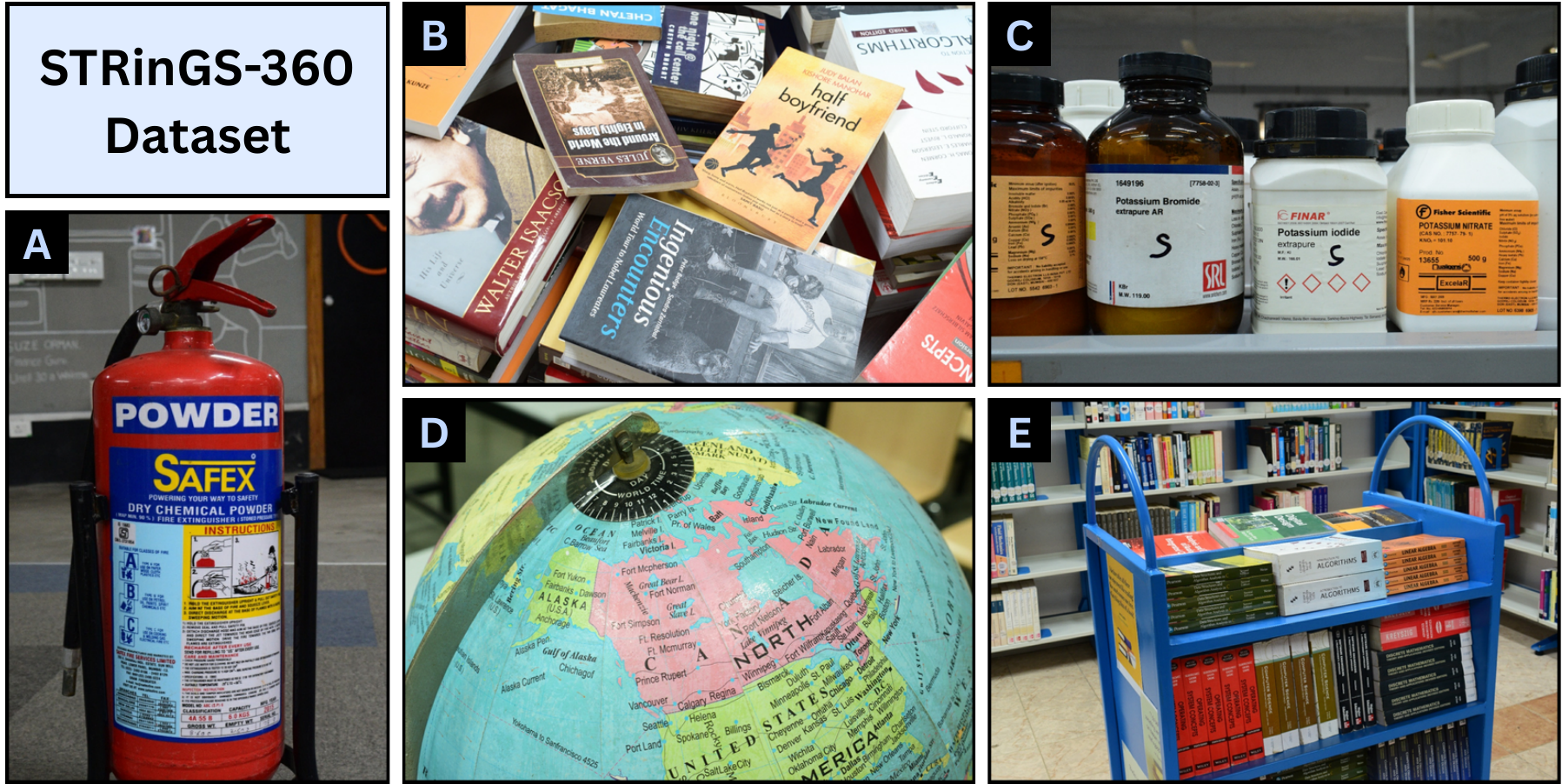}
\vspace{-6mm}
\captionof{figure}{Overview of the scenes in our STRinGS-360 dataset. Each scene contains semantically meaningful text elements:
(A)~Extinguisher,
(B)~Books,
(C)~Chemicals,
(D)~Globe, and
(E)~Shelf.
The dataset is designed to evaluate text reconstruction performance under diverse layouts and text orientations.}
\label{fig:dataset}
\vspace{-4mm}
\end{figure}

%% file: sec/4_methodology.tex
\section{\methodname{} Methodology}
\label{sec:method}

\input{figures/method}

We present an overview of \methodname{} in \cref{fig:method}.
We begin with preprocessing: SfM and text segmentation (\cref{subsec:preprocessing}) followed by segmenting text regions in 3D (\cref{subsec:localization}).
Next, we propose our two-phase optimization that selectively refines text regions (\cref{subsec:phase1}) followed by integration with non-text regions and full scene optimization (\cref{subsec:phase2}).

\input{algorithms/localization}

\subsection{Preprocessing}
\label{subsec:preprocessing}

\paragraph{COLMAP SfM.}
Given $n$ input images $\mathcal{I} = \{I_1, \ldots, I_n\}$ of a static scene captured from different viewpoints, 3DGS begins by extracting geometric information required for initialization.
Specifically, we obtain a sparse 3D point cloud of $m$ points $\mathcal{P} = \{\mathbf{P}_1, \ldots, \mathbf{P}_m\}$,
camera poses associated with the images $\mathcal{C} = \{C_1, \ldots, C_n\}$, and
the camera intrinsics $K$ using the COLMAP pipeline~\cite{schoenberger2016sfm, schoenberger2016mvs}.
Additionally, for each point $\mathbf{P}_i$, COLMAP provides a visibility set $V_i \subseteq \{1, \ldots, n\}$ indexing the subset of images in which the point is observed.
We denote the collection of these visibility sets as $\mcV = \{V_1, \ldots, V_m\}$.

\paragraph{Text segmentation.}
To identify and isolate textual regions in the undistorted images output by COLMAP, we employ Hi-SAM~\cite{ye2024hi}, a  model capable of segmenting text at multiple scales and orientations.
We refer to the binary mask for image $I_j$ as $M_j$, and the set of all masks as $\mcM = \{M_1, \ldots, M_n\}$.

\subsection{Text Segmentation in 3D}
\label{subsec:localization}

To enable text-aware reconstruction in our pipeline, we first identify the subset of 3D points that correspond to text regions in the scene.
This is done by projecting each 3D point (from COLMAP) into all images where it is visible, and checking whether its 2D projection falls inside the corresponding Hi-SAM text mask.
A point is classified as a text point if it lies within the text region in at least $\tau$ images.
In our method, we set the visibility threshold $\tau = 1$.
The set of \textit{text points} is denoted as $\Ptext \subseteq \mcP$, and its complement as $\Pnontext = \mcP \setminus \Ptext$.
The pseudo-code for this process is provided in \cref{alg:localization}.

The Gaussians used in 3DGS are initialized directly from the sparse point cloud $\mcP$, with each point providing the 3D location $(x, y, z)$ of a Gaussian.
Leveraging the text/non-text partitioning from above, we define $\Gtext$ and $\Gnontext$ as the initial sets of Gaussians corresponding to $\Ptext$ and $\Pnontext$ respectively.
These subsets serve as the basis of our two-phase training strategy described next.

\subsection{Phase 1: Selective Text Reconstruction}
\label{subsec:phase1}

We start GS training using the text Gaussians $\Gtext$, obtained through the 3D text segmentation process above.
This phase runs for $T_1$ iterations (3K), and optimization is performed on the subset of images with non-empty text masks.

\paragraph{Densification of text Gaussians.}
Since the initialization is based on a sparse point cloud, high-frequency structures (text) may be underrepresented, especially in cases where the number of viewpoints observing the text is small.
To address this, we adopt a visibility-based densification strategy at the start of phase 1.
Note, this is a one-time densification in addition to the standard densification process used in 3DGS.
Specifically, the number of duplicates $N_i$ for each Gaussian $\bfg_i \in \Gtext$ is inversely proportional to its visibility:
\begin{equation}
\small{
N_i = \left\lfloor \frac{1/c_i - \min_k(1/c_k)}
{\max_k(1/c_k) - \min_k(1/c_k)} \cdot 
(\Nmax {-} 1) + 1 \right\rceil.
}
\end{equation}
$c_i = |V_i|$ is the visibility count of point $\bfP_i$ and corresponding Gaussian $\bfg_i$.
The parameter $\Nmax$ defines the maximum densification factor, chosen to be between 15-25 based on the density of text in the scene.

We apply the densify-and-split strategy to each Gaussian, guided by its densification factor.
This results in multiple smaller Gaussians at slightly perturbed positions that cover the same volume, thereby enabling an efficient representation of text.
The result of this process is an augmented set of text Gaussians, denoted as $\Gtext^{\text{dense}}$.
The necessity and effectiveness of this densification are discussed in \cref{subsec:ablations}.

\paragraph{Text region loss.}
To ensure that the optimization focuses on text regions, we use the segmented text masks to modify the loss function.
Specifically, for an image $I_j$ and its rendered counterpart $R_j$, the reconstruction loss is:
\begin{equation}
\small{
\mcL_1^{\text{text}} = \| I_j \odot M_j - R_j \odot M_j \|_1 \, .
}
\end{equation}
where $\odot$ denotes element-wise multiplication and $M_j$ is the binary text mask.
This replaces the standard photometric loss formulation in 3DGS that combines $\mcL_1$ and D-SSIM terms over the entire image~\cite{kerbl20233d}.

\paragraph{Locking position parameters.}
3DGS typically employs a coarse-to-fine optimization schedule where the position parameters of Gaussians are updated with relatively high learning rates (LRs) at the start.
This often causes them to drift away from high-frequency regions such as text.
As our text Gaussians are initialized at text regions, we lock them in position by setting their position LR to zero, while allowing other parameters to be updated.

The output of phase 1 is a refined set of text Gaussians, denoted $\Gtextrefined$, used in phase 2 for full scene optimization.

\input{figures/lr_curves}

\subsection{Phase 2: Full Scene Refinement}
\label{subsec:phase2}
We now focus on jointly optimizing both text and non-text regions of the scene.
The refined text Gaussians $\Gtextrefined$ are combined with initial non-text Gaussians $\Gnontext$ obtained from 3D text segmentation process.
After $T_1$ (3K) iterations of phase 1, phase 2 runs up to $T_2$ (30K) iterations.

In this phase of training, we maintain the same loss function as in 3DGS~\cite{kerbl20233d}, including the D-SSIM component, to ensure full scene refinement. 
We also follow the standard procedures for densification, splitting, and cloning Gaussians as 3DGS.

\paragraph{Modulating position learning rates.}
A key concern is preserving the quality of $\Gtextrefined$ that may drift from their position if updated indiscriminately to minimize global photometric loss.
To address this, we apply a text region dependent LR for the positions of text and non-text Gaussians separately.

For $\Gtext$, we propose an \textit{increasing LR factor} as a sigmoid function.
This results in conservative early updates that preserve existing structure while providing flexibility later.
Conversely, for $\Gnontext$, we apply a constant multiplier $\alpha$ to ensure compatibility with the lowered LR for $\Gtext$ and avoid destabilizing updates.

\input{tables/ocr_cer_summary}

The region-specific LR factor $\eta_r(t)$ for the position of a Gaussian $\bfg$ at iteration $t \in [T_1, T_2]$ is:
\begin{equation}
\label{eq:pos_lr}
\small{
\eta_r(t) = 
\begin{cases}
\displaystyle \frac{\alpha}{1 + e^{-\beta \cdot (t - \gamma)}}
  & \text{if } \bfg \in \Gtextrefined \, , \\
\alpha & \text{if } \bfg \in \Gnontext \, .
\end{cases}
}
\end{equation}

Next, let $\eta_{\text{base}}(t)$ be the LR schedule adopted by vanilla 3DGS.
We shift this by $T_1$ iterations to obtain $\eta_{\text{opt}}(t)$.
Then, the effective LR used to update the position of each Gaussian is
$\eta_{\text{effective}}(\bfg, t) = \eta_r(t) \cdot \eta_{\text{opt}}(t)$,
and is illustrated in \cref{fig:lr_curves}. We explain hyperparameter choices in Appendix A.

Overall, \methodname{}'s hybrid strategy enables targeted and region-aware optimization, ensuring sharp and readable text while preserving overall scene quality.

%% file: figures/method.tex
\begin{figure*}[t]
\centering
\includegraphics[width=\linewidth]{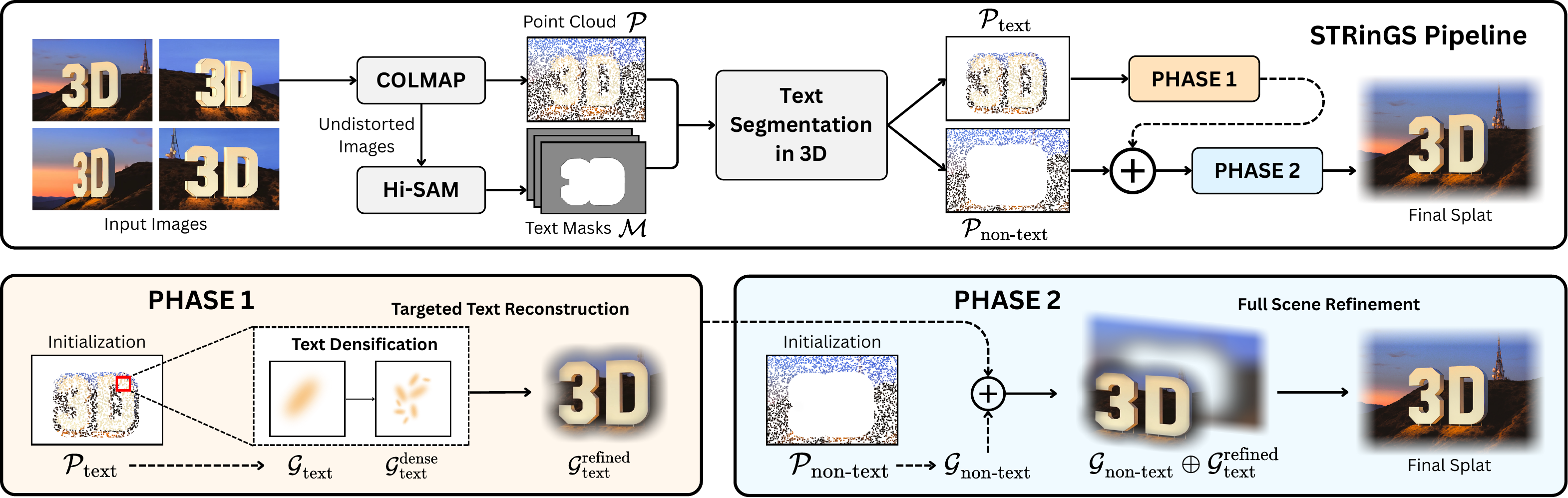}
\vspace{-6mm}
\captionof{figure}{\textbf{\methodname{} overview}.
Given $n$ input images, we use COLMAP to obtain a point cloud $\mcP$ and undistorted images, which are passed to Hi-SAM~\cite{ye2024hi} to obtain text masks $\mcM$.
$\mcP$ and $\mcM$ are passed to the Text Segmentation in 3D module (\cref{subsec:localization}, \cref{alg:localization}) to obtain partitioned text and non-text point clouds.
These are processed through a two-phase pipeline.
In phase 1 (\cref{subsec:phase1}), we perform targeted densification and reconstruction of text Gaussians.
In phase 2 (\cref{subsec:phase2}), we perform full scene refinement, where text and non-text Gaussians are optimized with distinct learning strategies, enabling targeted enhancement of text without compromising scene quality.
The final output is a text-refined Gaussian Splat representation with enhanced text readability while preserving overall scene fidelity.
}
\vspace{-3mm}
\label{fig:method}
\end{figure*}

%% file: algorithms/localization.tex
\begin{algorithm}[t]
\caption{Text Segmentation in 3D}
\label{alg:localization}
\KwIn{
Point cloud $\mathcal{P}$; camera intrinsics $K$; camera poses $\mathcal{C}$; text masks $\mathcal{M}$; visibility sets $\mathcal{V}$; visibility threshold $\tau$
}
\KwOut{$\Ptext, \Pnontext$}

$\Ptext \gets \emptyset$\

\For{each point $\mathbf{P}_i \in \mathcal{P}$, where $i = 1$ to $m$}{
    $\textit{count} \gets 0$\
    
    \For{each image index $j \in V_i$}{
        \tcp{Perspective Projection}
        $\mathbf{u}_{ij} \gets \pi(K, C_j, \mathbf{P}_i)$\
        
        \If{$M_j(\mathbf{u}_{ij}) = 1$}{
            $\textit{count} \gets \textit{count} + 1$\
        }
    }
    \If{$\textit{count} \geq \tau$}{
        $\Ptext \gets \Ptext \cup \{\mathbf{P}_i\}$\
    }
}

$\Pnontext \gets \mathcal{P} \setminus \Ptext$\

\Return{$\Ptext, \Pnontext$}
\end{algorithm}

%% file: figures/lr_curves.tex
\begin{figure}[b]
\centering
\vspace{-4mm}
\includegraphics[width=0.49\linewidth]{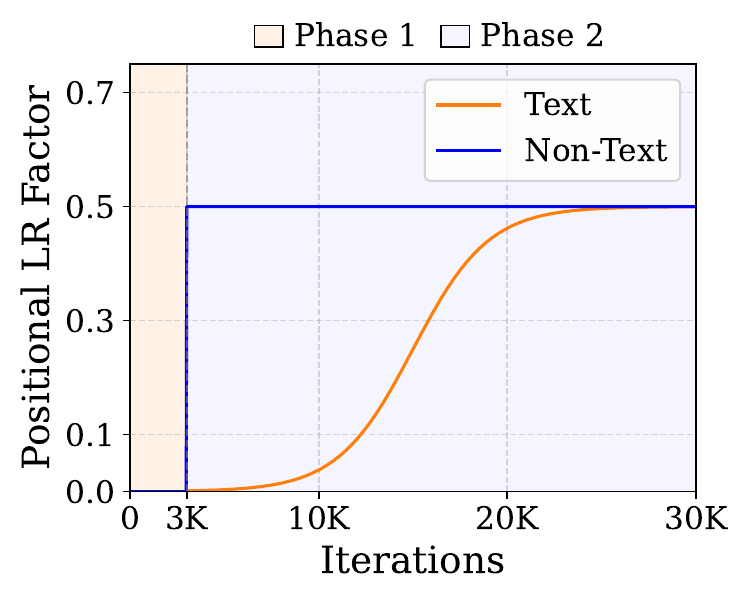}
\hfill
\includegraphics[width=0.49\linewidth]{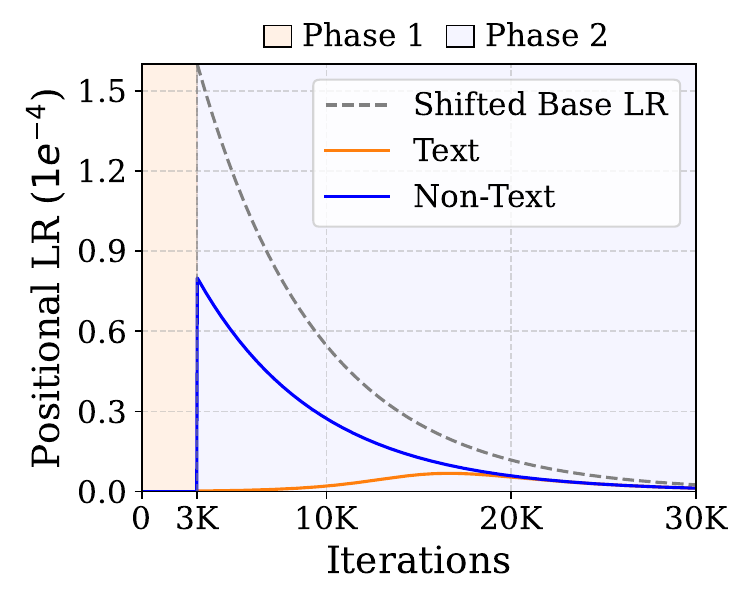}
\vspace{-3mm}
\caption{Learning rate (LR) of the position parameter for Gaussians in \methodname{} (see \cref{eq:pos_lr}).
\textbf{Left:} Learning rate scaling factor $\eta_r(t)$ for text and non-text Gaussians.
\textbf{Right:} Effective LR obtained by modulating a shifted base exponential decay schedule $\eta_\text{opt}(t)$ from 3DGS with these factors.
$\alpha{=}0.5$, $\beta{=}0.0005$, $\gamma{=}15000$.
Note, phase 1 sets the position learning rate of $\Gtext$ to 0 while $\Gnontext$ is not optimized. In phase 2, we introduce differentiated learning for text and non-text content.
}
\label{fig:lr_curves}
\end{figure}

%% file: tables/ocr_cer_summary.tex
\newcolumntype{C}[1]{>{\centering\arraybackslash}m{#1}} 

\begin{table}[t]
\centering
\footnotesize
\tabcolsep=0.8mm
\begin{tabular}{l *{6}{C{0.75cm}}}
\toprule    
\multirow{2}{*}{OCR-CER $\downarrow$} 
& \multicolumn{2}{c}{\makecell{TandT}} 
& \multicolumn{2}{c}{\makecell{DL3DV-10K}} 
& \multicolumn{2}{c}{\makecell{STRinGS-360}} \\
\cmidrule(lr){2-3} \cmidrule(lr){4-5} \cmidrule(lr){6-7}
& 7K & 30K & 7K & 30K & 7K & 30K \\
\midrule
3DGS~\papertable{SIGGRAPH'23}      & \cellcolor{third}0.209 & 0.121 & \cellcolor{third}0.392 & 0.157 & \cellcolor{third}0.736 & 0.148 \\
Mip-Splatting~\papertable{CVPR'24}      & 0.222 & 0.125 & \cellcolor{third}0.392 & \cellcolor{third}0.149 & 0.748 & 0.129 \\
3DGS-MCMC~\papertable{NeurIPS'24}  & 0.272 & \cellcolor{third}0.120 & 0.511 & \cellcolor{second}0.142 & 0.927 & \cellcolor{second}0.110 \\
AbsGS~\papertable{ACMMM'24}        & 0.249 & 0.137 & 0.411 & 0.160 & 0.768 & 0.143 \\
EDC-AbsGS~\papertable{arXiv'25}    & \cellcolor{second}0.142 & \cellcolor{second}0.118 & \cellcolor{second}0.239 & 0.162 & \cellcolor{second}0.328 & \cellcolor{third}0.116 \\
\textbf{STRinGS (Ours)}            & \cellcolor{first}0.122 & \cellcolor{first}0.099 & \cellcolor{first}0.187 & \cellcolor{first}0.123 & \cellcolor{first}0.177 & \cellcolor{first}0.106 \\
\bottomrule
\end{tabular}
\vspace{-2mm}
\caption{OCR-based Character Error Rate (CER $\downarrow$) on rendered images at 7K and 30K training iterations averaged over all scenes in the dataset. Lower CER indicates better text readability.
Red, orange, and yellow highlights indicate the \hlfirst{first}, \hlsecond{second}, and \hlthird{third} best performing technique.}
\label{tab:ocr_cer}
\vspace{-3mm}
\end{table}

%% file: sec/5_exp_and_results.tex
\section{Experiments and Results}
\label{sec:experiments_and_results}

\input{tables/training_time_summary}

\input{figures/comparison_7k}

Following standard protocol in the 3DGS literature~\cite{kerbl20233d}, every 8\textsuperscript{th} image is held out as an evaluation view to assess novel view synthesis performance.
Each scene is trained for $T_2$ (30K) iterations.
To evaluate results on early text reconstruction, we also report results at 7K iterations.
All experiments are conducted on an Nvidia RTX 3090 Ti GPU with 24GB VRAM.

The pipeline involves running COLMAP to obtain the sparse point cloud, camera poses, and undistorted images.
The undistorted images are passed to the Hi-SAM-L~\cite{ye2024hi} model which outputs tight polygonal text masks.
These are dilated using a circular kernel with a diameter equal to 5\% of the image width, thereby spanning the visual footprint of a text region, which includes the text strokes and immediate background context.
This is followed by the two-stage training procedure outlined in \cref{sec:method}.

\subsection{OCR-based Evaluation}
\label{subsec:ocr_based_evaluation}

3D reconstruction quality is typically measured using image-based metrics such as PSNR, SSIM, and LPIPS~\cite{zhang2018unreasonable}, which quantify similarity between rendered and ground-truth images.
They are computed by averaging pixel-level or perceptual differences over entire images, often dominated by background non-textual regions.
While effective at assessing global appearance, these metrics fall short in evaluating the semantic fidelity of reconstructed text.

In our scenes, even if text occupies a small fraction of the images, it has high semantic importance. 
Character-level distortions, misalignments, or partial blurring may severely impair text legibility, however, barely affects PSNR or SSIM scores.
To address this limitation, we introduce an OCR-based evaluation score that measures the quality of text reconstruction.
Specifically, we run Google OCR API~\cite{googlevisionapi} on the rendered views and the corresponding ground-truth images.
For each evaluation image, we compute the Character Error Rate (CER): the normalized Levenshtein distance between recognized and ground-truth text, using a recall-based approach that penalizes missing and mismatched ground-truth characters.
OCR-CER reflects how well the reconstructed image retains readable and accurate textual information. The CER scores are aggregated across all evaluation views within each scene.
Additional details are provided in Appendix C.

\subsection{Comparison with Existing Works}
\label{subsec:comparison_with_existing_works}

\paragraph{Baselines.}
We compare against
vanilla 3DGS~\cite{kerbl20233d} and other recent methods.
While there are no existing methods targeting text reconstruction, Mip-Splatting~\cite{yu2024mip},
3DGS-MCMC~\cite{kheradmand20243d}, AbsGS\cite{ye2024absgs}, and EDC-AbsGS~\cite{deng2024efficient} \footnote{By EDC-AbsGS, we refer to this implementation \url{https://github.com/XiaoBin2001/EDC} linked in their arXiv preprint.}
serve as strong baselines as they refine the overall scene.

\paragraph{Datasets.}
We evaluate all methods on a diverse set of 14 scenes drawn from existing benchmarks and \dataname{}.
This includes
2 scenes from the Tanks and Temples dataset~\cite{knapitsch2017tanks},
7 selected scenes from the DL3DV-10k Benchmark~\cite{ling2024dl3dv} that feature varying amounts of textual content, and
5 scenes from our \dataname{} dataset, consisting of sharp, dense, and semantically meaningful text.

\input{figures/comparison_30k}

\paragraph{Text reconstruction results.}
We compare model performance at two stages:
7K and 30K iterations.
\cref{tab:ocr_cer} shows that \methodname{} achieves the lowest OCR-CER, with a big gap at 7K iterations.
The relative improvements, averaged over all datasets are: 63.6\% 3DGS, 64.3\% Mip-Splatting, 71.6\% 3DGS-MCMC, 66.0\% AbsGS, and 31.4\% EDC-AbsGS.
\cref{fig:comparison_with_existing_works_7k} visualizes the noticeably sharper and readable text at 7K iterations for various scenes.
\methodname{} does especially well on reconstructing small text such as
``acetaminophen'' (row 2), ``product code 18060'' (row 3), or names on the globe such as ``Minneapolis'' (row 4).

While other methods bridge the gap at 30K iterations, \methodname{} still outperforms them with a relative improvement in OCR-CER scores: 23.0\% 3DGS, 18.6\% Mip-Splatting, 11.8\% 3DGS-MCMC, 25.4\% AbsGS, and 17.2\% EDC-AbsGS.
A few examples are visualized in \cref{fig:comparison_with_existing_works_30k}.

\methodname{} is most effective when text regions contain few points at initialization or when the text is visible in a small subset ($<$ 5\%) of images, where other methods tend to fail.
Importantly, this targeted text refinement results in comparable overall scene quality and fewer Gaussians (\cref{tab:metrics_30k}).
What distinguishes our method from others is its ability to accurately reconstruct small text, whereas other methods can already handle large text reasonably well, as detailed in Appendix D.2.
We also demonstrate the effectiveness of our method on multilingual text refinement in Appendix D.1.

\subsection{Ablations and Key Highlights}
\label{subsec:ablations}

\paragraph{Effect of text densification.}
To address sparse points at initialization in text regions leading to under reconstruction, we introduce a targeted text densification step in phase 1 (\cref{subsec:phase1}).
As illustrated in \cref{fig:densification_ablation}, the benefits of text-aware densification are evident.
Vanilla 3DGS fails to reconstruct the text even after 30K iterations, while \methodname{} without text densification also fails to reconstruct the text.
Our approach with densification successfully reconstructs sharp and accurate text at 30K iterations while clearly showing a few letters even at earlier stages of training. Results showing the effect of text densification are presented in \cref{tab:ablations}.
We see consistent improvements in OCR-CER indicating better text reconstruction across all datasets.

\input{tables/ablations}

\paragraph{Effect of position LR of Gaussians.}
To assess the impact of the position LR during phase 1 (\cref{subsec:phase1}), we evaluate the outputs at the end of this phase using OCR-CER.
As shown in \cref{fig:lr_ablation}, using a non-zero LR for the positions of Gaussians leads to significant degradation in text reconstruction.
This is especially important in our setting, where Gaussians are already densely placed over text regions through explicit densification.
By setting the position LR to zero, we freeze their locations, allowing the optimization of other parameters such as scale, opacity, and spherical harmonic coefficients leading to sharper text reconstruction. Results in \cref{tab:ablations} show that zero position LR is crucial for improving text quality from the early stages (3K iterations of phase 1) indicated by the significantly improved OCR-CER.

\input{figures/densification_ablation}

\input{tables/metrics_30k}

\input{figures/text_evolution}

\paragraph{Training speed.}
Our method achieves better text reconstruction quality with significantly lower training time, compared to existing densification-based approaches (\cref{tab:training_time}). 
Densification in standard 3DGS relies on large positional gradients to dynamically add Gaussians during training, which introduces significant computational overhead. In contrast, \methodname{} sets the position LR to zero in the first phase and keeps it lower than 3DGS in the second, effectively limiting unnecessary densification. Since we explicitly add Gaussians in text regions at the start of phase 1, we avoid the need for extensive gradient-driven densification, leading to faster and more efficient training.

While EDC-AbsGS is the strongest baseline in terms of CER, compared to \methodname{}, it requires 2.8$\times$ training time for 7K iterations and 1.7$\times$ for 30K iterations. On the other hand, 3DGS is closest to \methodname{} in training time (only 1.4$\times$ at both 7K and 30K), but performs significantly worse in text reconstruction quality (\cref{tab:ocr_cer}). These results highlight that \methodname{} performs the best in terms of both efficiency and accuracy. The trade-off between OCR-CER and training time across methods is visualized in \cref{fig:teaser}.
A detailed breakdown of the time required for preprocessing (COLMAP and text segmentation) and training (phases 1 and 2) is provided in Appendix D.3.

\paragraph{Early text reconstruction.}
We demonstrate the evolution of text reconstruction quality over training iterations on the \textit{Extinguisher} scene from our dataset.
Our method achieves noticeably better text reconstruction at early stages (3K and 7K iterations) compared to vanilla 3DGS (\cref{fig:text_evolution}). 
The accompanying plot illustrates the evolution of OCR-CER across iterations, showing that our method reconstructs text accurately much earlier.

\input{figures/position_lr_ablation}

\subsection{Discussion}
\label{sec:discussions}

\paragraph{Applications.}
\methodname{} is well-suited for use cases where both quality and efficiency are critical. For example, autonomous navigation requires early recovery of readable text for tasks like interpreting signs/directions and waypoint recognition. In robotics, clear reconstruction of text assists in scene understanding and labeled object identification.
In AR/VR environments, user experience is enhanced by good quality of reconstructed text. Further, \methodname{} may prove valuable in cultural heritage applications, where reconstructing inscriptions such as ancient stone carvings, temple wall engravings, or historical monument plaques as 3D models can aid archival and restoration efforts.

\paragraph{Limitations.}
\methodname{} uses Hi-SAM for 2D text segmentation that introduces computational overhead during preprocessing and may miss text in cluttered scenes. However, this can be swapped out for future models that improve text segmentation.
Future work could focus on reducing Hi-SAM’s computational overhead, for instance by performing 3D text segmentation on only a strategically chosen subset of images rather than the full set. 
Additionally, \methodname{} fails when text in input images is unreadable due to low resolution, making reconstruction inherently limited.

%% file: tables/training_time_summary.tex
\begin{table}[t]
\centering
\footnotesize
\tabcolsep=0.8mm
\begin{tabular}{l *{6}{C{0.75cm}}}
\toprule    
Training Time
& \multicolumn{2}{c}{\makecell{TandT}} 
& \multicolumn{2}{c}{\makecell{DL3DV-10K}} 
& \multicolumn{2}{c}{\makecell{STRinGS-360}} \\
\cmidrule(lr){2-3} \cmidrule(lr){4-5} \cmidrule(lr){6-7}
(in minutes) $\downarrow$ 
& 7K & 30K & 7K & 30K & 7K & 30K \\
\midrule
3DGS~\papertable{SIGGRAPH'23}     & \cellcolor{second}2.0 & 13.8 & \cellcolor{second}2.8 & \cellcolor{second}15.1 & \cellcolor{second}2.5 & \cellcolor{second}17.2 \\
Mip-Splatting~\papertable{CVPR'24}      & 3.4 & 20.8 & 6.3 & 30.4 & 5.7 & 31.7 \\
3DGS-MCMC~\papertable{NeurIPS'24} & 3.1 & 19.6 & \cellcolor{third}5.2 & 28.3 & 5.6 & 34.0 \\
AbsGS~\papertable{ACMMM'24}       & \cellcolor{third}2.6 & \cellcolor{second}12.7 & 5.3 & \cellcolor{third}20.7 & \cellcolor{third}5.4 & \cellcolor{third}22.5 \\
EDC-AbsGS~\papertable{arXiv'25}   & 2.8 & \cellcolor{second}12.7 & 6.0 & 22.2 & 5.5 & 23.2 \\
\textbf{STRinGS (Ours)}           & \cellcolor{first}1.1 & \cellcolor{first}9.6 & \cellcolor{first}2.1 & \cellcolor{first}11.4 & \cellcolor{first}1.9 & \cellcolor{first}12.6 \\
\bottomrule
\end{tabular}
\vspace{-2mm}
\caption{Training time in minutes at 7K and 30K training iterations, averaged over all scenes in the dataset.}
\label{tab:training_time}
\vspace{-5mm}
\end{table}

%% file: figures/comparison_7k.tex
\begin{figure*}[t]
\centering
{\small
\begin{tabularx}{\linewidth}{*{6}{>{\centering\arraybackslash}X}}
3DGS~\cite{kerbl20233d} & Mip-Splatting~\cite{yu2024mip} & 3DGS-MCMC~\cite{kheradmand20243d} & EDC-AbsGS~\cite{deng2024efficient} & \textbf{STRinGS (Ours)} & Ground Truth \\
\end{tabularx}}
\includegraphics[width=\linewidth]{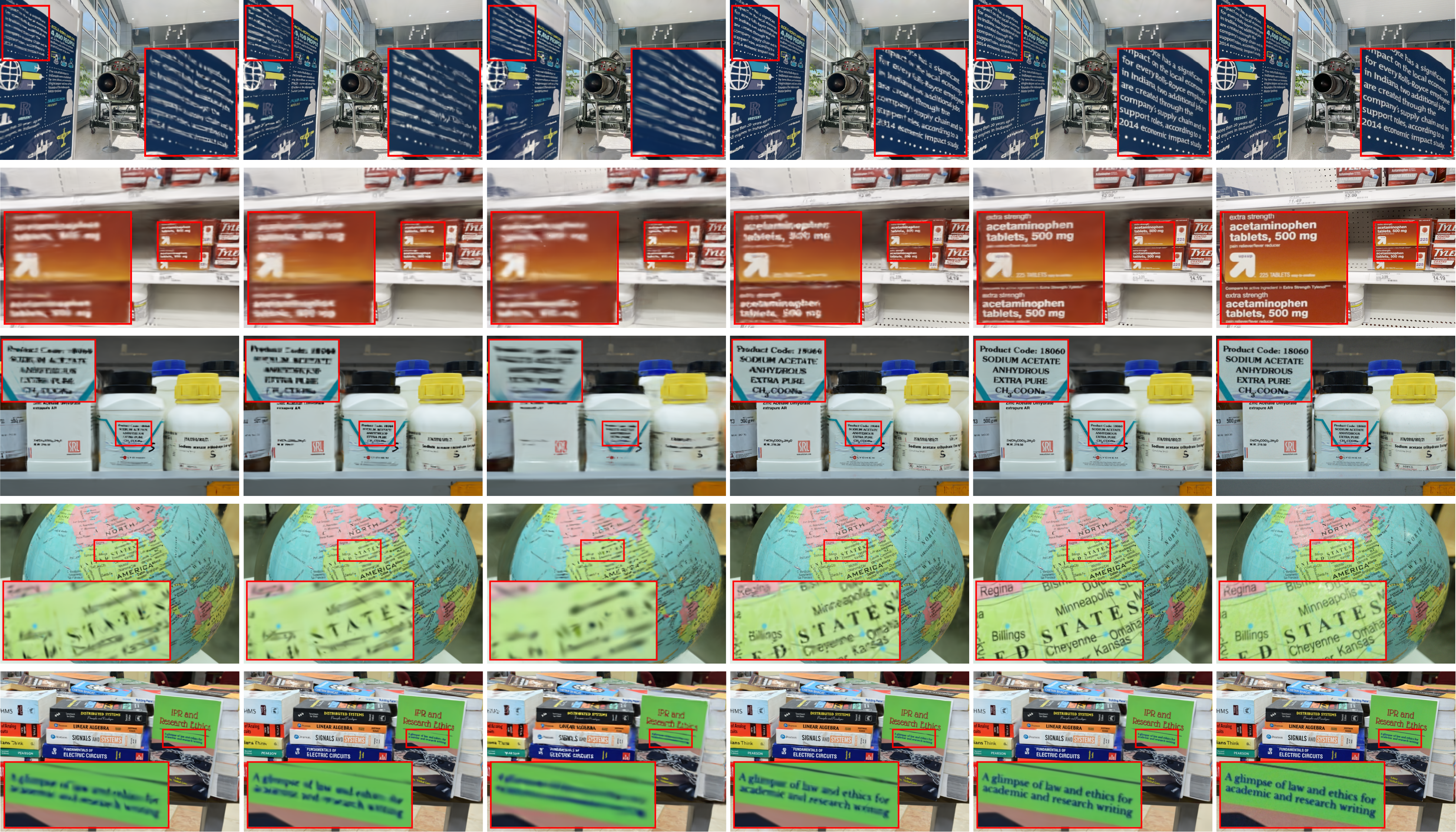}
\vspace{-6mm}
\captionof{figure}{Qualitative comparison of different methods at 7K training iterations on scenes from the DL3DV-10K Benchmark~\cite{ling2024dl3dv} (rows 1, 2) and our \dataname{} (rows 3-5) datasets.
While existing methods struggle to reconstruct text accurately at this early stage, our \methodname{} framework produces significantly sharper and more legible text regions.
(Best seen on screen)}
\vspace{-3mm}
\label{fig:comparison_with_existing_works_7k}
\end{figure*}

%% file: figures/comparison_30k.tex
\begin{figure*}[t]
\centering
\small
\begin{tabularx}{\linewidth}{*{6}{>{\centering\arraybackslash}X}}
3DGS~\cite{kerbl20233d} & Mip-Splatting~\cite{yu2024mip} & 3DGS-MCMC~\cite{kheradmand20243d} & EDC-AbsGS~\cite{deng2024efficient} & \textbf{STRinGS (Ours)} & Ground Truth \\
\end{tabularx}
\includegraphics[width=\linewidth]{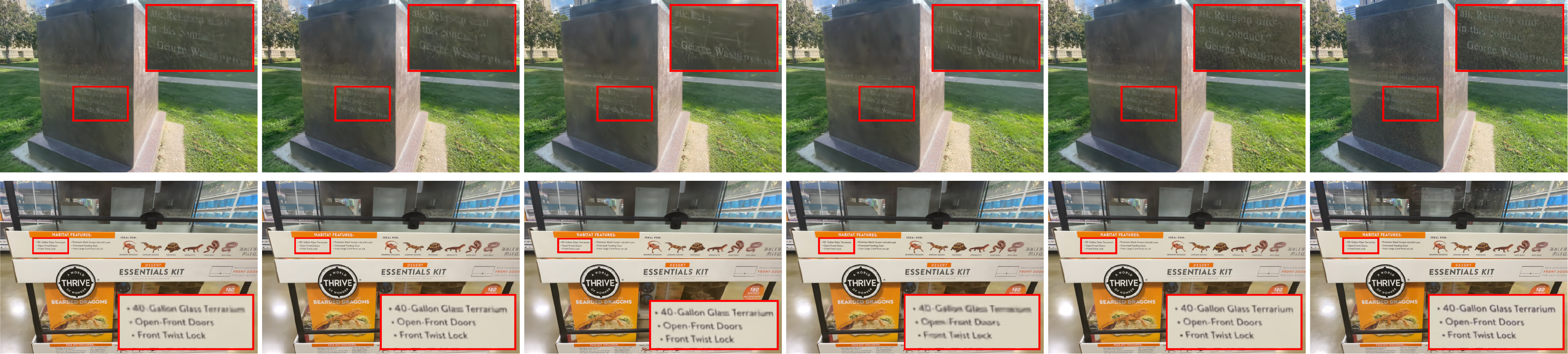}
\vspace{-6mm}
\captionof{figure}{Qualitative comparison of different methods at 30K training iterations on scenes from DL3DV-10K Benchmark~\cite{ling2024dl3dv}.
\methodname{} consistently preserves text clarity, even in visually challenging regions where other methods miss fine textual details.
(Best seen on screen)}
\label{fig:comparison_with_existing_works_30k}
\vspace{-3mm}
\end{figure*}

%% file: tables/ablations.tex
\begin{table}[b]
\vspace{-4mm}
\centering
\footnotesize
\tabcolsep=1mm
\begin{tabular}{C{2.4cm} *{6}{C{0.7cm}}}
\toprule
Dataset & \multicolumn{2}{c}{TandT} & \multicolumn{2}{c}{DL3DV-10K} & \multicolumn{2}{c}{STRinGS-360} \\
\cmidrule(lr){1-7}
& \multicolumn{6}{c}{\cellcolor{gray}{Effect of text densification}} \\
\multirow{2}{*}{\shortstack{OCR-CER $\downarrow$\\(7K iterations)}} & w/o & Ours & w/o & Ours & w/o & Ours \\
& 0.196 & \textbf{0.122} & 0.316 & \textbf{0.187} & 0.437 & \textbf{0.177} \\
\cmidrule(lr){1-7}
& \multicolumn{6}{c}{\cellcolor{gray}{Effect of zero position LR of Gaussians}} \\
\multirow{2}{*}{\shortstack{OCR-CER $\downarrow$\\(3K iterations)}} & w/o & Ours & w/o & Ours & w/o & Ours \\
& 0.342 & \textbf{0.289} & 0.618 & \textbf{0.278} & 0.948 & \textbf{0.347}\\
\bottomrule
\end{tabular}
\vspace{-2mm}
\caption{Ablations. The effect of text densification and the effect of zero position LR of Gaussians in phase 1. The OCR-CER values, averaged over all scenes in the datasets demonstrate the necessity of both components for accurate text reconstruction.}
\label{tab:ablations}
\end{table}

%% file: figures/densification_ablation.tex
\begin{figure}[b]
\vspace{-5mm}
\centering
\includegraphics[width=\linewidth]{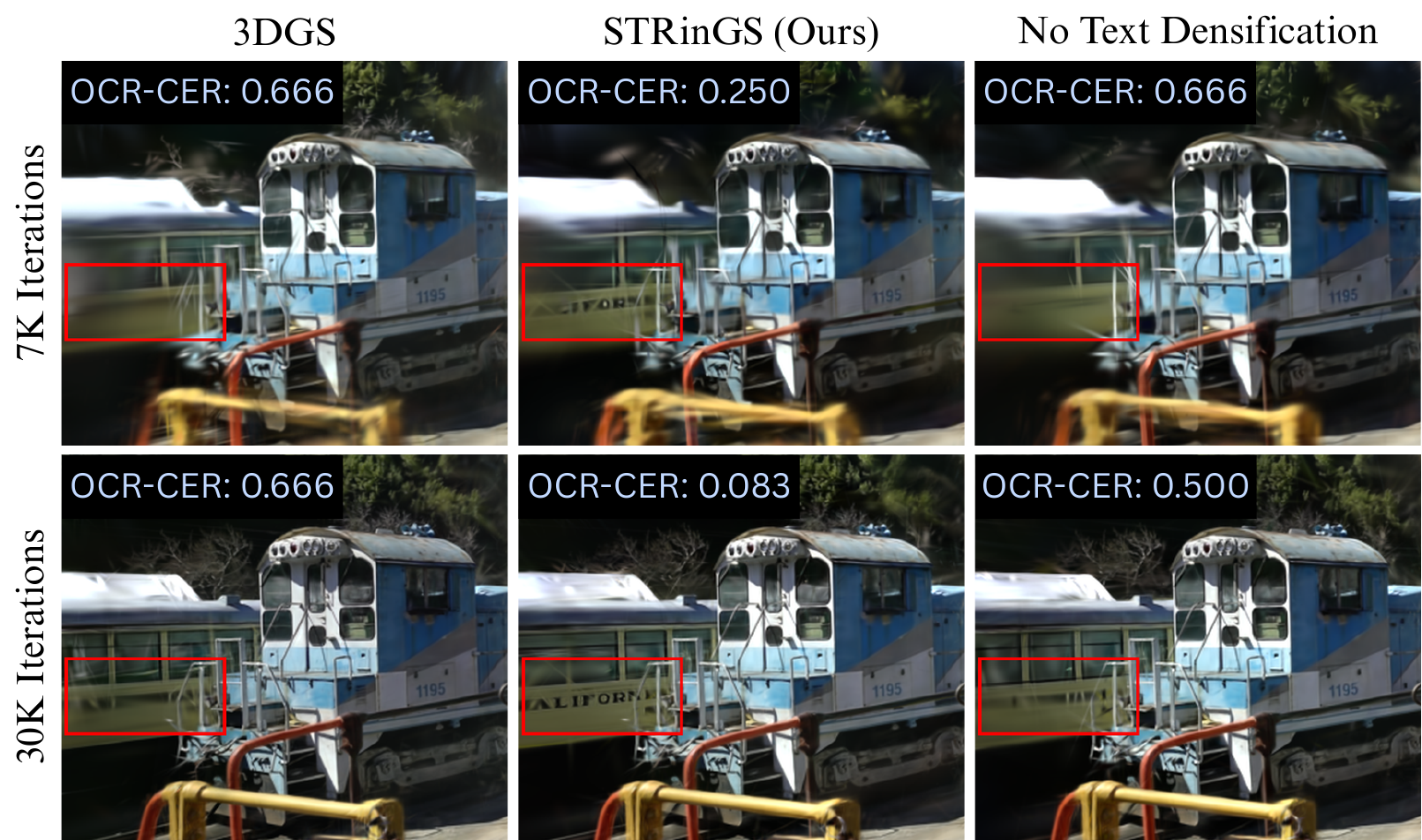}
\vspace{-6mm}
\captionof{figure}{Effect of text densification on a scene from the Tanks\&Temples~\cite{knapitsch2017tanks} dataset. 
\textbf{Left}: Vanilla 3DGS fails to reconstruct readable text even after 30k iterations, resulting in high OCR-CER of 0.666.
\textbf{Middle}: \methodname{} (Ours) with text densification achieves sharp and semantically meaningful text as early as 7K iterations which improves further at 30K iterations (0.083 CER).
\textbf{Right}: Without text densification, our method struggles to produce accurate and legible text, demonstrating the importance of targeted densification of text regions.}
\vspace{-3mm}
\label{fig:densification_ablation}
\end{figure}

%% file: tables/metrics_30k.tex
\begin{table*}[h]
\centering
\footnotesize
\renewcommand{\arraystretch}{1.1}
\setlength{\tabcolsep}{5pt} 

\begin{tabular}{lcccccccccccc}
\toprule
\multirow{2}{*}{Method} 
& \multicolumn{4}{c}{Tanks\&Temples} 
& \multicolumn{4}{c}{DL3DV-10K Benchmark}
& \multicolumn{4}{c}{STRinGS-360 (Ours)} \\ 
\cmidrule(lr){2-5} \cmidrule(lr){6-9} \cmidrule(lr){10-13}
& PSNR$\uparrow$ & SSIM$\uparrow$ & LPIPS$\downarrow$ & Points$\downarrow$
& PSNR$\uparrow$ & SSIM$\uparrow$ & LPIPS$\downarrow$ & Points$\downarrow$
& PSNR$\uparrow$ & SSIM$\uparrow$ & LPIPS$\downarrow$ & Points$\downarrow$ \\
\midrule
3DGS~\papertable{SIGGRAPH'23}      & 23.73 & 0.8524 & 0.1692 & 1576K & 30.20 & 0.9348 & 0.1456 & 1175K & 28.85& 0.9126& 0.2107& 1391K \\

Mip-Splatting~\papertable{CVPR'24}     & \cellcolor{third}23.81& \cellcolor{first}0.8596& \cellcolor{third}0.1563& 2366K& \cellcolor{first}30.47& \cellcolor{second}0.9390& \cellcolor{second}0.1329& 1610K& 28.80& \cellcolor{third}0.9142& \cellcolor{third}0.2012& 1875K\\

3DGS-MCMC~\papertable{NeurIPS'24}     & \cellcolor{first}24.43 & 0.7688 & \cellcolor{first}0.1508 & 1550K & \cellcolor{second}30.46 & \cellcolor{second}0.9390 & 0.1394 & 1182K & \cellcolor{first}29.85& \cellcolor{first}0.9234& \cellcolor{first}0.1971& 1388K\\

AbsGS~\papertable{ACMMM'24}        & 23.64 & \cellcolor{third}0.8526 & 0.1616 & \cellcolor{first}1297K & 30.18 & 0.9360 & \cellcolor{third}0.1368 & \cellcolor{second}874K & 28.77& 0.9111& 0.2044& \cellcolor{third}1240K\\

EDC-AbsGS~\papertable{arXiv'25}       & 23.73 & \cellcolor{second}0.8595 & \cellcolor{second}0.1557 & \cellcolor{third}1382K & \cellcolor{third}30.45 & \cellcolor{first}0.9400 & \cellcolor{first}0.1321 & \cellcolor{first}857K & \cellcolor{second}29.30& \cellcolor{second}0.9183& \cellcolor{second}0.1992& \cellcolor{second}1041K\\

\textbf{STRinGS (Ours)}& \cellcolor{second}23.88 & 0.8513 & 0.1767 & \cellcolor{second}1354K & 30.14 & 0.9338 & 0.1477 & \cellcolor{third}918K & \cellcolor{third}29.00& 0.9138& 0.2166& \cellcolor{first}965K\\

\bottomrule
\end{tabular}

\vspace{-2mm}
\caption{Comparison of reconstruction quality and number of Gaussians (Points) at 30K iterations across three datasets: Tanks\&Temples~\cite{knapitsch2017tanks}, DL3DV-10K~\cite{ling2024dl3dv}, and STRinGS-360. Our method achieves comparable PSNR, SSIM, and LPIPS scores, indicating no degradation in overall scene quality, while requiring slightly lesser Points especially in text-rich scenes (STRinGS-360 dataset).}
\label{tab:metrics_30k}
\vspace{-3mm}
\end{table*}

%% file: figures/text_evolution.tex
\begin{figure*}[t]
\vspace{2mm}
\centering
\includegraphics[width=\linewidth]{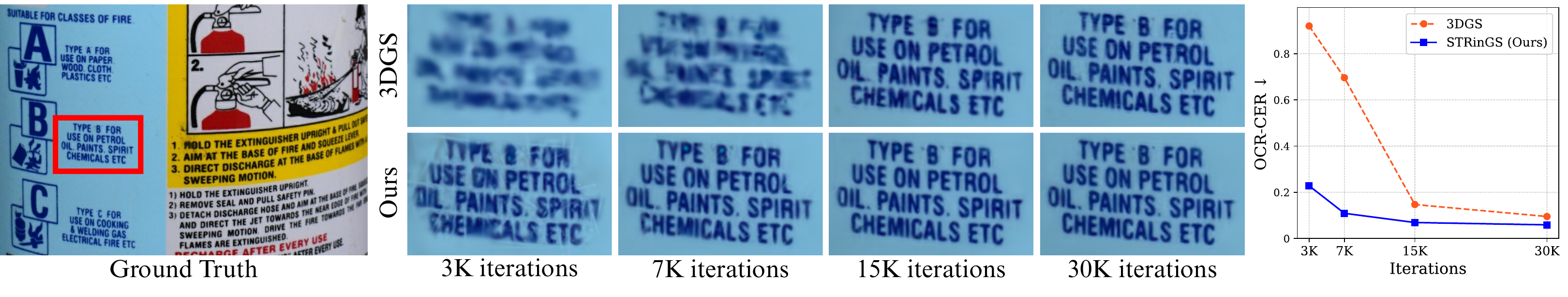}
\vspace{-6mm}
\captionof{figure}{Text reconstruction across training iterations on the \textit{Extinguisher} scene from our STRinGS-360 dataset. STRinGS achieves clearer and more accurate text reconstruction earlier than 3DGS, as reflected in the plot for OCR-CER of the scene over iterations.}
\label{fig:text_evolution}
\vspace{-4mm}
\end{figure*}

%% file: figures/position_lr_ablation.tex
\begin{figure}[b]
\vspace{-2mm}
\centering
\includegraphics[width=\linewidth]{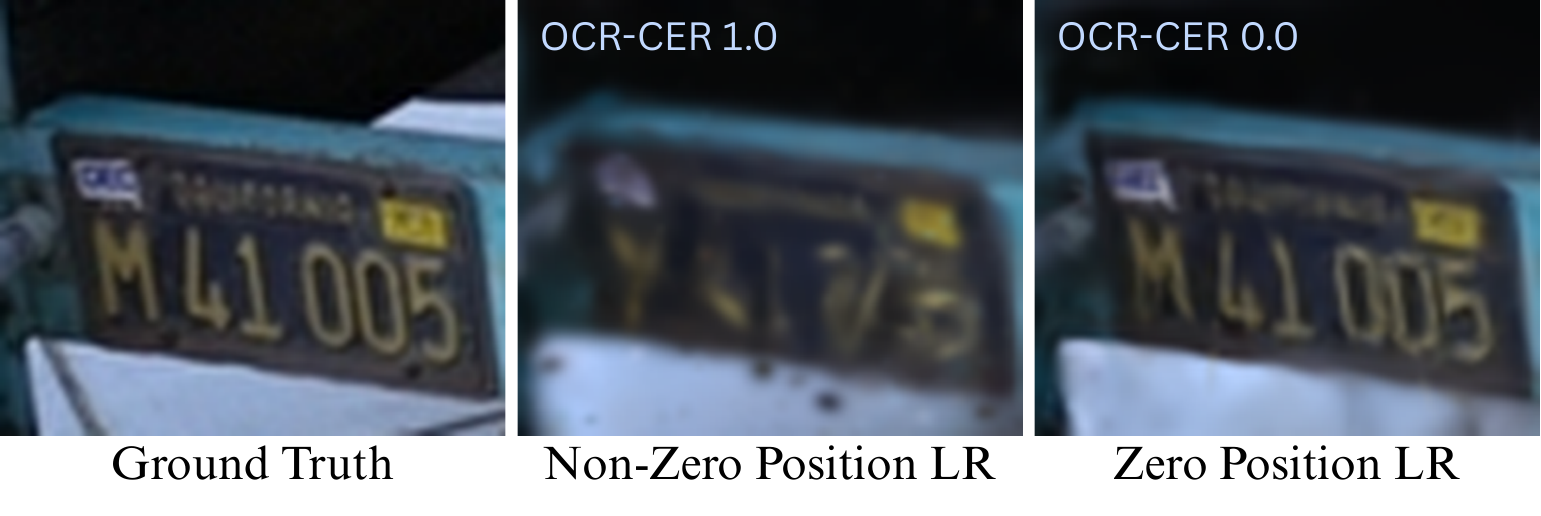}
\vspace{-6mm}
\captionof{figure}{Effect of position learning rate at the end of phase 1 (3K iterations) on a scene from the Tanks\&Temples~\cite{knapitsch2017tanks} dataset.
A non-zero LR causes Gaussians to drift, leading to poor text reconstruction (CER = 1.0).
Instead, freezing their positions (zero LR) preserves spatial alignment, enabling text readability (CER = 0.0).}
\vspace{-3mm}
\label{fig:lr_ablation}
\end{figure}

%% file: sec/6_conclusion.tex
\section{Conclusion}
\label{sec:conclusion}
We introduced \methodname{}, a novel text-aware refinement framework that explicitly focuses on reconstructing sharp, clear and readable text. By treating text and non-text regions separately, our two-phase optimization enables early recovery of textual content. 
Extensive evaluations across diverse text-rich scenes demonstrated that \methodname{} consistently outperforms baselines, achieving significantly lower OCR-based Character Error Rates, particularly at early iterations, highlighting its potential for time-sensitive applications.
We also proposed \dataname{}, a curated dataset specifically designed for evaluating text readability in 3D reconstructions. By using OCR-CER as a measure for text readability, we quantitatively validated the improvements offered by our method over vanilla 3DGS and its variants.
In summary, \methodname{} establishes a new direction for text-aware 3D scene understanding, highlighting the importance of semantic detail preservation in 3D scene reconstruction.

{\small
\paragraph{Acknowledgements.}
We thank Harshavardhan P. for feedback and guidance throughout this project.
RKS thanks Digital India Bhashini Division, Ministry of Electronics and Information Technology (MeiTY), Government of India for supporting the project, and MT thanks Adobe Research for travel support.
}